\documentclass[10pt,twocolumn,letterpaper]{article}

\usepackage{iccv}
\usepackage{times}
\usepackage{epsfig}
\usepackage{graphicx}
\usepackage{amsmath}
\usepackage{amssymb}
\usepackage{subcaption}
\PassOptionsToPackage{hyphens}{url}\usepackage[pagebackref=true,breaklinks=true,letterpaper=true,colorlinks,bookmarks=false]{hyperref}

\iccvfinalcopy 

\def\iccvPaperID{33} 

\ificcvfinal\fi

\begin{document}

\title{Exploiting Temporality for Semi-Supervised Video Segmentation}

\author{Radu Sibechi \thanks{Work done during an internship at TomTom.}\\
University of Amsterdam\\
{\tt\small radu.sibechi@gmail.com}
\and
Olaf Booij \qquad Nora Baka\\
TomTom, Amsterdam\\
{\tt\small firstname.lastname@tomtom.com}
\and
Peter Bloem \\
VU Amsterdam\\
{\tt\small vu@peterbloem.nl}
}

\maketitle
\ificcvfinal\thispagestyle{empty}\fi

\begin{abstract}
    In recent years, there has been remarkable progress in supervised image segmentation. Video segmentation is less explored, despite the temporal dimension being highly informative. Semantic labels, \eg that cannot be accurately detected in the current frame, may be inferred by incorporating information from previous frames. However, video segmentation is challenging due to the amount of data that needs to be processed and, more importantly, the cost involved in obtaining ground truth annotations for each frame. In this paper, we tackle the issue of label scarcity by using consecutive frames of a video, where only one frame is annotated. We propose a deep, end-to-end trainable model which leverages temporal information in order to make use of easy to acquire unlabeled data. Our network architecture relies on a novel interconnection of two components: a fully convolutional network to model spatial information and temporal units that are employed at intermediate levels of the convolutional network in order to propagate information through time. The main contribution of this work is the guidance of the temporal signal through the network. We show that only placing a temporal module between the encoder and decoder is suboptimal (baseline). Our extensive experiments on the CityScapes dataset indicate that the resulting model can leverage unlabeled temporal frames and significantly outperform both the frame-by-frame image segmentation and the baseline approach. 
\end{abstract}

\section{Introduction}

\begin{figure}[h]
\begin{center}
\includegraphics[width=\linewidth]{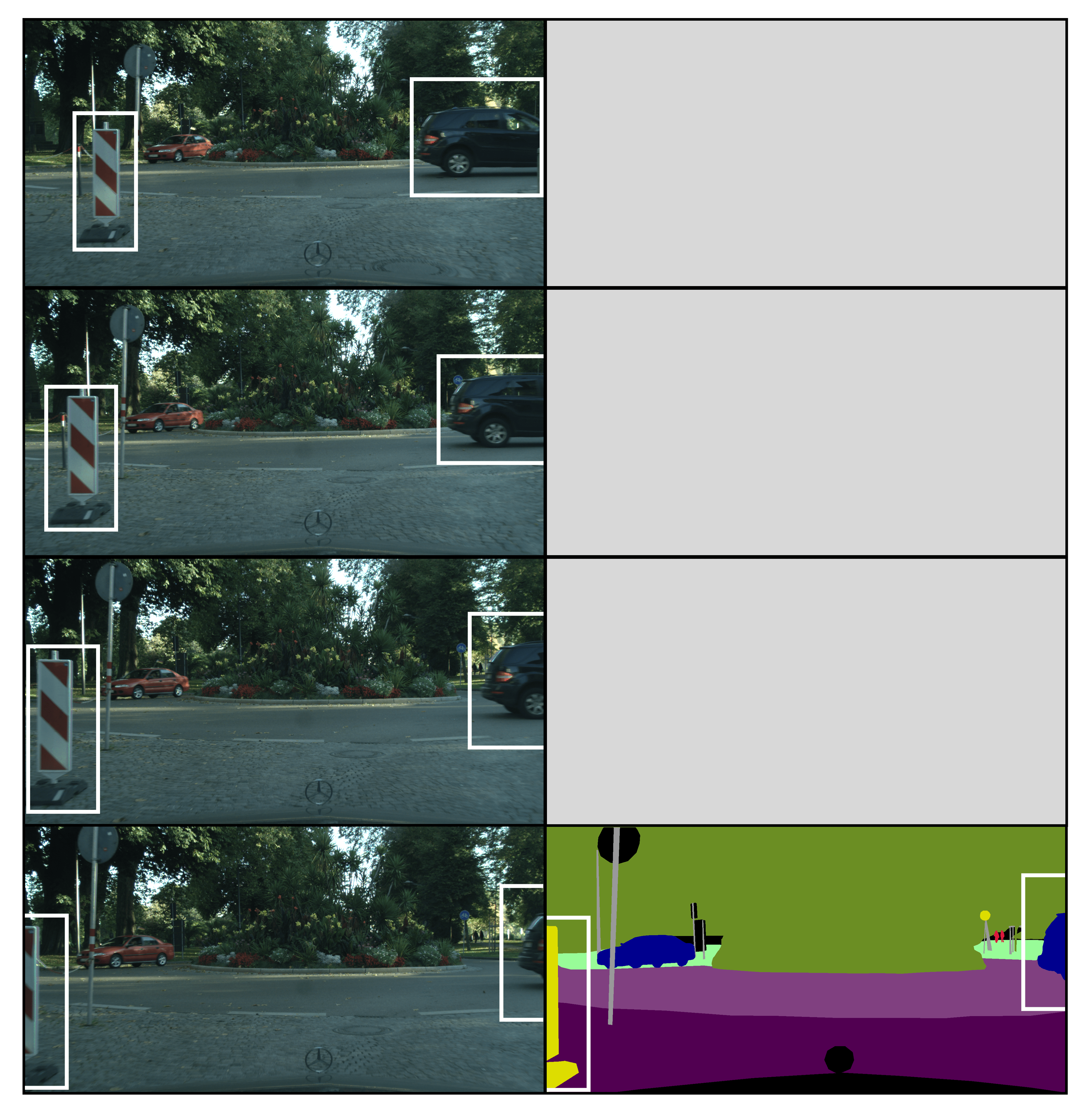}
\end{center}
   \caption{Illustration of a case in which temporal information is highly beneficial. In the current frame (bottom), both the sign on the left and the car on the right are mostly occluded and cannot be accurately classified. By including unlabeled information from previous frames, we can infer their type and propagate this information through time, in order to correctly classify them in the present.}
\label{sequence-example}
\end{figure}

Visual understanding of complex scenes is an essential component of advanced real-world systems. A particularly popular and challenging application involves self-driving cars, which make extreme demands on system performance and reliability. There has been remarkable progress in this area and many sophisticated methods based on deep neural networks have been proposed \cite{DBLP:journals/pami/ShelhamerLD17,DBLP:journals/corr/RonnebergerFB15,DBLP:journals/corr/abs-1802-02611,DBLP:journals/corr/ZhaoSQWJ16}. A major contributing factor to their success is the availability of large-scale, densely annotated, public datasets. When it comes to semantic segmentation, data for training and refining single frame models is now quite diverse \cite{DBLP:journals/corr/CordtsORREBFRS16,DBLP:journals/corr/LinMBHPRDZ14}. In contrast, obtaining detailed annotations for individual video frames is more time consuming and costly, which makes frame-level supervision for video segmentation inherently difficult.

Since videos are  sequences of images, one could segment individual images independently, ignoring the time dimension. However, the time dimension is a rich source of information, and is required for accurate predictions in some cases. For example, in Figure~\ref{sequence-example}, we show that objects which are mostly occluded in the current frame, can be inferred by including information from previous frames. Videos provide additional information such as long-range temporal interactions among objects, casual relations among events and motion of objects in the scene. Based on these observations, one can propagate information through time in order to leverage temporal dependencies. The question is how to make image segmentation models suitable for handling the spatio-temporal dimension. The key challenge is effectively exploiting the information that is available in the temporal dimension when ground truth annotations are scarce and frame-level supervision is impossible. 

We propose a spatio-temporal deep neural network for semantic video segmentation by using consecutive frames from a video. The proposed network architecture combines a lightweight fully-convolutional U-Net architecture \cite{DBLP:journals/corr/RonnebergerFB15} with temporal units that are employed at intermediate levels of the convolutional network in order to propagate spatial information through time.

The main contributions of our work are the following:
\begin{itemize}
    \item We present a novel, deep, end-to-end trainable architecture for semantic video segmentation that leverages the temporal dimension in order to propagate spatial information. 
    \item We propose a lightweight module for transforming traditional, fully convolutional networks into spatio-temporal FCNs. Specifically, without substantially increasing the model complexity, the proposed method can be easily added in already published state-of-the-art methods.
    \item We show that the model is able to correctly classify objects that cannot be accurately detected in the current frame by inferring them from previous frames. By incorporating cheap, unlabeled, temporal data we are able to significantly outperform the frame-by-frame baseline on the CityScapes dataset.
    \item We show that current state of the art approaches which model temporality between the encoder and the decoder are suboptimal for capturing motion information.
\end{itemize}

\section{Related work}
State of the art approaches for semantic image segmentation are based on fully-convolutional networks (FCNs) \cite{DBLP:journals/corr/abs-1802-02611, DBLP:journals/corr/ZhaoSQWJ16,DBLP:journals/corr/RonnebergerFB15}. In U-Net \cite{DBLP:journals/corr/RonnebergerFB15}, the authors propose a symmetric network in which the encoder gradually reduces the feature maps and captures higher semantic information while the decoder module gradually recovers the spatial information. Skip connections through concatenation are used in order to exchange information between the encoder and the decoder. In order to capture contextual information at multiple scales, models such as PSPNet \cite{DBLP:journals/corr/ZhaoSQWJ16} perform spatial pyramid pooling. DeepLabv3+ \cite{DBLP:journals/corr/abs-1802-02611} applies several parallel dilated convolution with different rates, while also making use of the encoder decoder paradigm in order to recover object boundaries

\begin{figure*}
\begin{center}
\begin{subfigure}[b]{0.28\textwidth}
    \includegraphics[width=\textwidth]{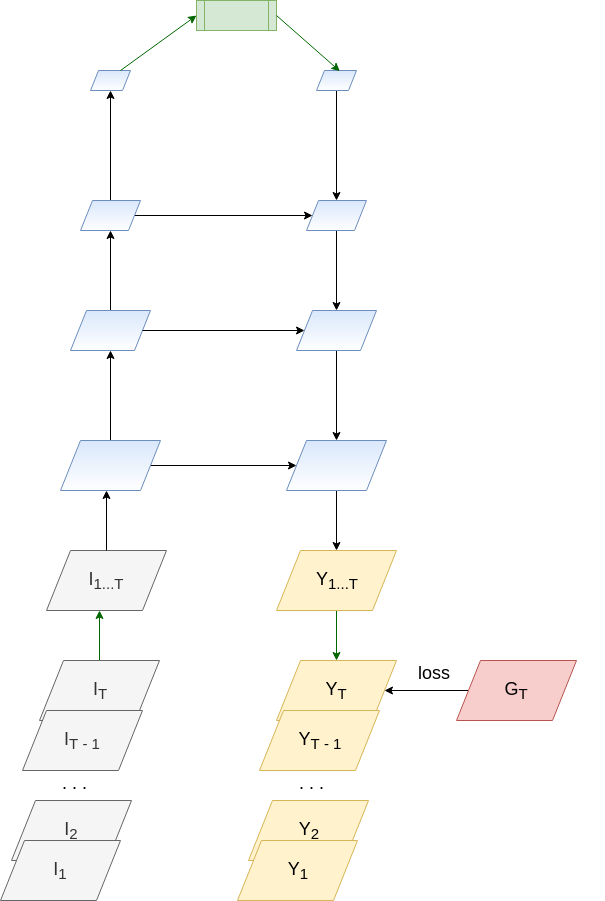}
    \caption{Modelling temporality between the encoder and decoder}
    \label{fig:temporal_bottleneck}
\end{subfigure}
\begin{subfigure}[b]{0.28\textwidth}
    \includegraphics[width=\textwidth]{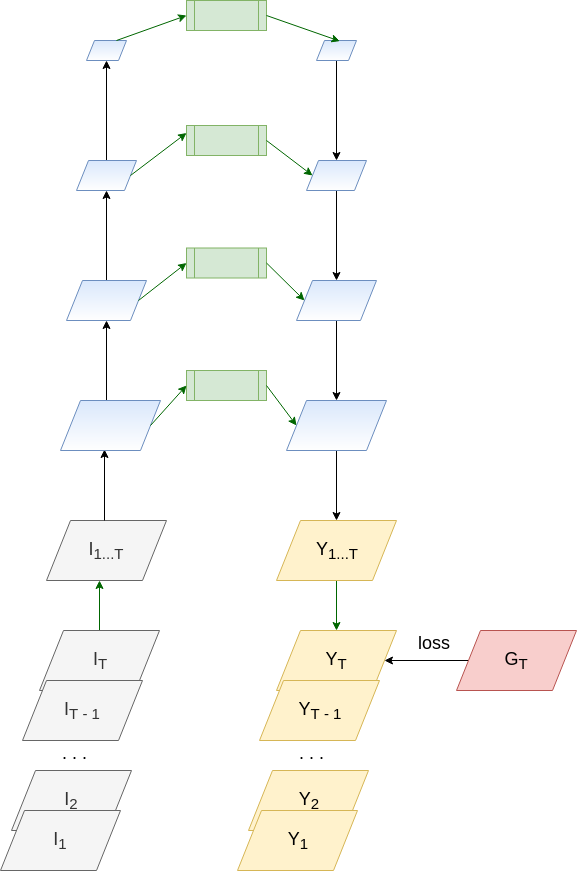}
    \caption{Modelling temporality at every skip connection level}
    \label{fig:temporal_skip}
\end{subfigure}
\begin{subfigure}[b]{0.28\textwidth}
    \includegraphics[width=\textwidth]{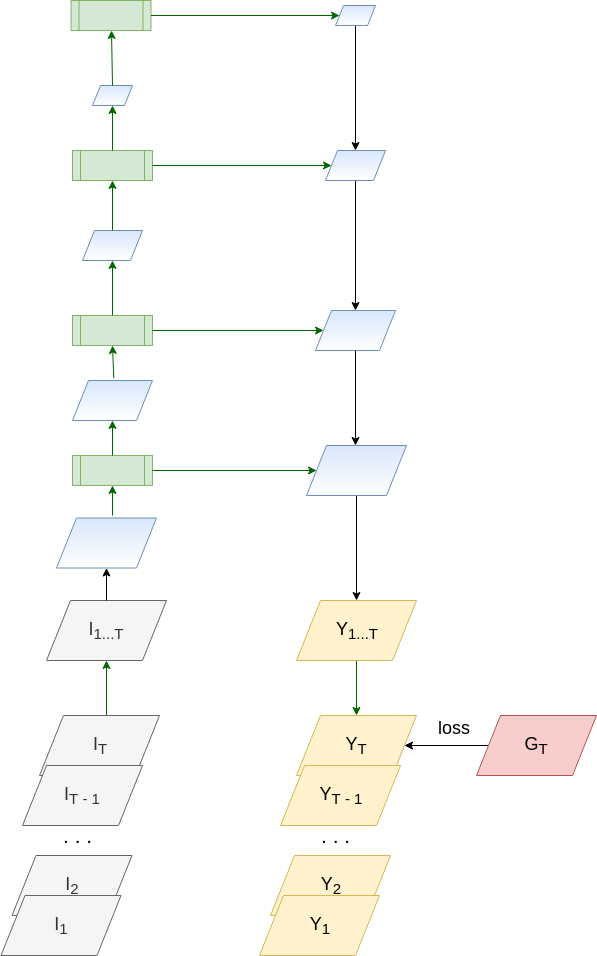}
    \caption{Proposed method - Propagating temporal features in the encoder}
    \label{fig:temporal_encoder}
\end{subfigure}
\begin{subfigure}[b]{0.12\textwidth}
    \includegraphics[width=\textwidth]{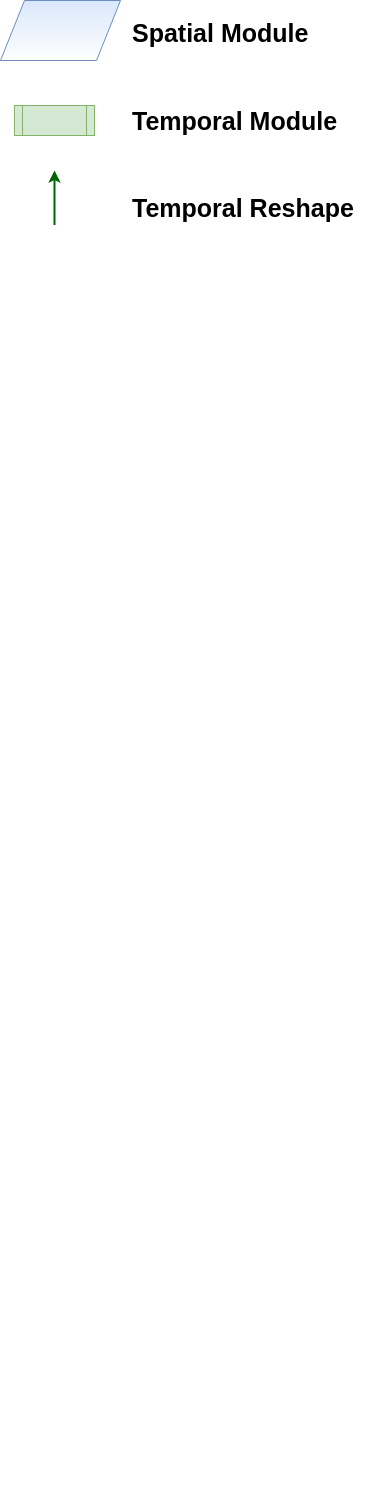}
\end{subfigure}
   \caption{Overview of different approaches to model temporality in U-Net based architectures.}
\label{overview-methods}
\end{center}
\end{figure*}

FCNs are not designed to model temporal dependencies. In order to bypass this problem, a naive solution is to concatenate multiple consecutive frames and make this the input of the FCN, resulting in an extra time dimension. Learning a representative set of 3D spatio-temporal convolution kernels is challenging and computationally intensive due to the high complexity of 3D kernels and the required amount of training videos. The performance gained by applying such a method is below 2\% on the Sports-1M benchmark, for the video classification task \cite{6165309}.  In order to extend FCNs to video data and avoid the added complexity of 3D kernels, we can split them into two components: one to model spatial information and one to model temporal information. Two-stream fully-convolutional networks fuse motion and spatial information in a unified framework \cite{DBLP:journals/corr/JainXG17a,DBLP:journals/corr/abs-1709-06750,DBLP:journals/corr/TokmakovAS17}. The hypothesis is that optical flow allows the model to retain the benefits of motion information, while still capturing global video information. Unfortunately, optical flow is often inaccurate, particularly around object boundaries. Another drawback is that they depend on optical flow models pretrained on different datasets, since large datasets with ground truth for both tasks do not exist. 

The combination of fully-convolutional networks with temporal recurrent units can solve many of the pitfalls of the previous approach. State of the art approaches first apply spatial filters by using a fully-convolutional encoder network, then a recurrent unit to model the temporality and finally use a decoder to get the desired output \cite{DBLP:journals/corr/NilssonS16,DBLP:journals/corr/FayyazSSFK16,DBLP:journals/corr/abs-1809-03327,DBLP:journals/corr/LeaFVRH16}. In \cite{DBLP:journals/corr/NilssonS16}, the authors describe an end-to-end architecture which combines a convolutional architecture and a spatio-temporal transformer layer that is able to propagate labeling information through time. In \cite{DBLP:journals/corr/FayyazSSFK16}, the spatio-temporal FCN is introduced, which uses a layer grid of Long-Short Term Memory (LSTM) models in between the FCN encoder and FCN decoder. In \cite{DBLP:journals/corr/abs-1809-03327}, the authors place a ConvLSTM \cite{DBLP:journals/corr/ShiCWYWW15} between the fully convolutional encoder and decoder for the task of video object segmentation, which concerns the segmentation of foreground objects from the background. For recognizing actions in videos, \cite{DBLP:journals/corr/LeaFVRH16} uses Temporal Convolutional Networks (TCNs) to process the features extracted by a pretrained FCN from multiple frames of a video. For the same task, \cite{ballas-2015-delving} learns spatio-temporal filters by stacking ConvGRUs at different locations in a pretrained deep convolutional neural network and iterating through video frames. In this direction, factorized spatio-temporal convolutional network (FSTCN) was proposed in \cite{DBLP:journals/corr/SunJYS15} for human action recognition. This network factorizes the standard 3D CNN model as a sequential process of learning 2D spatial kernels in the lower layers, followed by learning 1D temporal kernels in the upper layers.

Our work closely relates to \cite{ballas-2015-delving}, \cite{DBLP:journals/corr/LeaFVRH16}, \cite{DBLP:journals/corr/abs-1809-03327} and \cite{DBLP:journals/corr/SunJYS15}. Similar to \cite{ballas-2015-delving}, we also employ temporal units at every level of a FCN. However, we also propagate temporal features to subsequent convolutional layers. Furthermore, both \cite{ballas-2015-delving} and \cite{DBLP:journals/corr/LeaFVRH16} use a pre-trained, frozen CNN as a feature extractor, unlike our model, which is end-to-end trainable. Similar to \cite{DBLP:journals/corr/SunJYS15,DBLP:journals/corr/LeaFVRH16,DBLP:journals/corr/abs-1809-03327}, we also transform traditional fully convolutional networks into spatio-temporal convolutional networks. However, the architectures presented in \cite{DBLP:journals/corr/SunJYS15,DBLP:journals/corr/abs-1809-03327} use only one temporal module, in between a fully-convolutional encoder and a fully-connected decoder, respectively fully-convolutional decoder.  In \cite{DBLP:journals/corr/LeaFVRH16}, the authors stack multiple 1D convolutions to process the features extracted by a pre-trained frozen FCN from multiple video frames. These methods only learn temporal combinations for the features of only one convolutional layer. In our model, we learn different temporal combinations for every convolutional layer features and propagate them forward through the network to the next convolutional layers.

\section{Approach}

The purpose of this work is to adapt a fully-convolutional semantic segmentation network so that it can model the temporal dimension and be applied on video data. Given a sequence of $T$ video frames $I_1, I_2, ..., I_T$ the task is to predict semantic segmentations for each video frame. Ground truth annotations are only present for the last frame, $I_T$.

\subsection{Where to model temporality?}

In this section, we will discuss in detail different approaches for employing temporal units in an encoder-decoder network in order to propagate information through time. The visual illustration of how our proposed methods aggregate information from adjacent video frames is presented in Figure~\ref{overview-methods}.

\subsubsection{Frame-by-frame} We apply a U-Net-based fully-convolutional architecture~\cite{DBLP:journals/corr/RonnebergerFB15} for frame-by-frame image segmentation. When applying this method on a sequence of frames, each segmentation output is independent of previous frames. U-Net architectures follow the encoder-decoder architecture. Every step in the decoder consists of an upsampling of the feature map and a concatenation with the corresponding feature maps from the encoder (skip-connections). We train only on images $I_T$ that have ground truth $G_T$, resulting in a fully supervised approach for the frame-by-frame method. 
In our U-Net \cite{DBLP:journals/corr/RonnebergerFB15} implementation, a block of the encoder network includes one $3 \times 3$ convolutional layer, followed by a batch normalization layer \cite{DBLP:journals/corr/IoffeS15}, LeakyReLU non-linearity and maxpooling. Compared to the original paper, we use more blocks and reduce the number of convolutional layers in a block. This results in an accuracy/speed trade-off, where we halve the epoch training time with a minimal drop in segmentation accuracy. Reducing the training time is especially helpful for the temporal modules, where the amount of data that needs to be processed increases linearly with the number of annotated images. In the decoder, upsampling is done using strided transposed convolutions. We use the same normalization technique as the encoder, and ReLU for non-linearity \cite{Nair:2010:RLU:3104322.3104425}.

\subsubsection{Between the encoder and decoder (bottleneck)}
Similar to \cite{DBLP:journals/corr/SunJYS15,DBLP:journals/corr/abs-1809-03327}, temporality can be modeled between the encoder and decoder, the bottleneck. We feed the $T$ consecutive images $I_{1...T}$ as input to the FCN encoder in order to obtain features $F_{1...T}^{(L)}$ from the last encoder convolutional block $L$. These features $F_{1...T}^{(L)}$ are processed as a sequence by a temporal unit in order to model the temporal dimension and then sent to the decoder in order to obtain the segmentation prediction $Y_T$. An overview of this method can be seen in Figure~\ref{fig:temporal_bottleneck}.

\subsubsection{At every skip connection level}
Along the lines of \cite{ballas-2015-delving}, temporality can be modelled at every skip connection level. This way, the decoder sees how spatial features from every convolutional level change in time and is able to better make use of temporal information to produce the final segmentation. Formally, before concatenating the features $F^{(l)}_{1...T}$ extracted by the $l$\textsuperscript{th} encoder convolutional block, we apply the same temporal forward procedure as before. An overview of this method can be seen in Figure~\ref{fig:temporal_skip}. 

\subsubsection{Proposed method: Propagating temporal features in the encoder}
A drawback of the previous approaches is the fact that only the decoder uses features that include temporal information, through the use of skip connections and the bottleneck connection. Encoder features that are propagated forward do not include any temporal information and are independent of previous frames. To circumvent this, features $F_{1...T}^{(l)}$ resulted from applying a temporal forward procedure are sent as input to the next spatial convolutional block. This way, we ensure that the encoder benefits from the temporal connections as well. An overview of this method can be seen in Figure~\ref{fig:temporal_encoder}.

\subsection{How to model temporality?}

In this section, we propose several temporal modules, focusing on maintaining the segmentation performance of the network while decreasing the memory and training time overhead.

\subsubsection{ConvLSTM}
One example of a machine learning technique that learns temporal features is a Long Short-Term Memory Network (LSTM) \cite{Hochreiter:1997:LSM:1246443.1246450}. These have been combined with FCNs \cite{DBLP:journals/corr/VinyalsTBE14,DBLP:journals/corr/WangYMHHX16} in order to learn spatial and temporal filters. 3D feature maps of shape $C \times H \times W$ are flattened into 1D vectors of size $CHW$, where $C$ represents the channels, $H$ the height and $W$ the width of the feature maps. A disadvantage of this approach is that the data that flows through the LSTM is 1D, and as such we lose spatial information. Furthermore, as feature maps become bigger, the size of the vectorized features increases quadratically. One aproach that mitigates this issue and is suitable for dealing with sequence of images is the ConvLSTM \cite{DBLP:journals/corr/ShiCWYWW15}. It is a recurrent model, just like the LSTM, but internal matrix multiplications are exchanged with convolution operations. As a result, the data that flows through the ConvLSTM cells keep the input dimension, instead of being just 1D vectors of features.
The total number of parameters
is equal to $ 4C^2K^2$, where $K$ is the spatial kernel size, and is set to 3 in our experiments, independent of the input feature map shape.  

\subsubsection{Temporal Networks (TN)}
Although ConvLSTMs have shown very good performance at sequence modelling tasks, stacking multiple instances of them in a network is costly, both in training time and in memory requirements. Recent work \cite{DBLP:journals/corr/abs-1803-01271,DBLP:journals/corr/LeaFVRH16}, has shown that convolutions can be used for sequence modelling. The main advantage of using convolutions is that they have low memory requirements for training. Furthermore, unlike \mbox{ConvLSTM}, they are parallelizable, which results in faster training and inference. 

\par{\textbf{TN block.}} The architecture of our TN block is inspired from \cite{DBLP:journals/corr/abs-1803-01271}. Within a block, the TN has two layers of dilated convolution and ReLU non-linearity \cite{Nair:2010:RLU:3104322.3104425}. We also make use of the residual connection introduced in \cite{DBLP:journals/corr/HeZRS15}. The final output of a TN block is the output of the convolutions added to the input x of the block.

\par{\textbf{Pointwise TN.}} We adapt the TN block architecture, by making use of pointwise 1D convolutions (kernel size is 1). We treat the time dimension as the convolution channel dimension and convolve over the flattened 1D vector $CHW$. The main advantage of this approach is that regardless of feature map shape, we always need $T^2$ parameters for our temporal network. 

In the original TCN paper \cite{DBLP:journals/corr/abs-1803-01271}, the authors use $CHW$ as the convolutional channel dimension and convolve over $T$. Padding is used in order to maintain the output sequence length, which can result in artifacts, especially for short sequences. Furthermore, we would need $(CHW)^2K$ parameters, where $K$ is the temporal kernel size. It is obvious to see that stacking such temporal networks at the first convolutional layers is not possible, due to the amount of parameters. These are not issues for our modified Pointwise~TN.

\par{\textbf{2DHW TN.}} An issue with our previous Pointwise TN approach is the fact that we treat each pixel in each channel independently. We tackle this issue by using 2D convolutions in our TN block. The incoming feature maps of shape $T \times C \times H \times W$ are reshaped to $TC \times H \times W$. We treat the $TC$ dimension as the convolution channel dimension and convolve over $H \times W$. The total number of parameters is $T^2C^2K^2$, where the temporal kernel size $K$ is set to 2 in our experiments.

\section{Experiments} 
\subsection{Dataset}
For training the segmentation models, we use the CityScapes dataset \cite{DBLP:journals/corr/CordtsORREBFRS16}. The dataset consists of 5000 video sequences of high quality images (1024 $\times$ 2048 resolution), partitioned into 2975 train, 500 validation and 1525 test sequences. The videos are captured from 50 different cities in Germany and Switzerland, under different weather conditions. The authors have divided each video into 30 images and segmented each 20\textsuperscript{th} frame. In the \textit{conventional} version of the dataset we only have access to the video frames that have ground truth annotations. In order to make use of temporality, we use the \textit{sequence} version of the dataset, where we have access to all 30 frames, not just the annotated one. The original RGB frames and annotations are reshaped to $256 \times 512$ for memory and speed concern. For data augmentation, operations of random horizontal flip, random gaussian blur and color jitter are applied. The same data preprocessing is used for both frame-by-frame and sequence models. It is important to note that state of the art frame-by-frame segmentation models \cite{DBLP:journals/corr/abs-1802-02611,DBLP:journals/corr/ZhaoSQWJ16} pretrain their networks on the \textit{coarse} dataset, which contains an extra 20000 coarsely annotated images. This is not possible in our case, since no \textit{sequence} version exists for these extra images.

\subsection{Training}
For training the frame-by-frame model, we only use images that have ground truth. For the sequence models, we sample 4 consecutive images (14\textsuperscript{th}, 16\textsuperscript{th}, 18\textsuperscript{th} and 20\textsuperscript{th}) as the input to train the proposed networks. Based on the ground truth label on the 20\textsuperscript{th} frame, we can construct the training set. Cross entropy loss function is used to solve the discriminative segmentation task. We follow the standard protocol of using 19 semantic labels for evaluation without considering the void label. For the frames without ground truth annotations, loss is set to 0. To train the proposed networks efficiently we use the Adam optimizer \cite{Kingma2015AdamAM}, with the learning rate set to $1e^{-4}$. L2 regularization with weight decay rate of $5e^{-4}$ is used to avoid over-fitting. We also clip the gradient norm to not exceed the threshold of 5. We optimized these hyper-parameters for the frame-by-frame model and used the same values for the temporal models. The performance is measured in terms of mean pixel intersection-over-union (mIoU), averaged across multiple runs.

\begin{table}
\begin{center}
\caption{Quantitative results on the CityScapes validation set for the proposed methods, where we model temporality at the bottleneck, every skip connection level and in the encoder as well.}\label{results_ours}
\begin{tabular}{|l|l|}
\hline
Method &  mIoU class \\
\hline
U-Net \cite{DBLP:journals/corr/RonnebergerFB15} reimplementation & 0.572 $\pm$  0.0017  \\
U-Net ours & 0.563 $\pm$ 0.0012 \\
\hline
U-Net Pointwise TN bottleneck & 0.575 $\pm$ 0.0015  \\
U-Net Pointwise TN skip & 0.596 $\pm$ 0.0015  \\
U-Net Pointwise TN encoder & \textbf{0.614} $\pm$  0.0025 \\
\hline
U-Net 2DHW TN bottleneck & 0.582 $\pm$ 0.0016 \\
U-Net 2DHW TN skip & 0.603 $\pm$ 0.0015\\
U-Net 2DHW TN encoder & \textbf{0.622} $\pm$ 0.0017 \\
\hline
U-Net ConvLSTM bottleneck & 0.581 $\pm$ 0.0018  \\
U-Net ConvLSTM skip & 0.612 $\pm$ 0.0021  \\
U-Net ConvLSTM encoder & \textbf{0.631} $\pm$ 0.0024 \\
\hline
\end{tabular}
\end{center}
\end{table}

\begin{table}
\begin{center}
\caption{Quantitative results on the CityScapes test set for the proposed temporal encoder methods. We compare our method against other methods that perform inference at $256 \times 512$ resolution.}\label{results_test_ours}
\begin{tabular}{|l|l|l|}
\hline
Method &  mIoU class & \# params \\
\hline
U-Net frame-by-frame (ours) & 0.568 & 183M \\
U-Net Pointwise TN & 0.605 &  183M\\
U-Net 2DHW TN & 0.614 &  541M\\
U-Net ConvLSTM & \textbf{0.622} & 585M\\
\hline
Fast-SCNN \cite{DBLP:journals/corr/abs-1902-04502} & 0.519 & - \\
SegNet \cite{DBLP:journals/corr/BadrinarayananK15} & 0.561 & - \\
ENet \cite{DBLP:journals/corr/PaszkeCKC16} & 0.583 & -  \\
\hline
\end{tabular}
\end{center}
\vspace{-4mm}
\end{table}

\begin{figure*}
\begin{center}
\includegraphics[width=\linewidth]{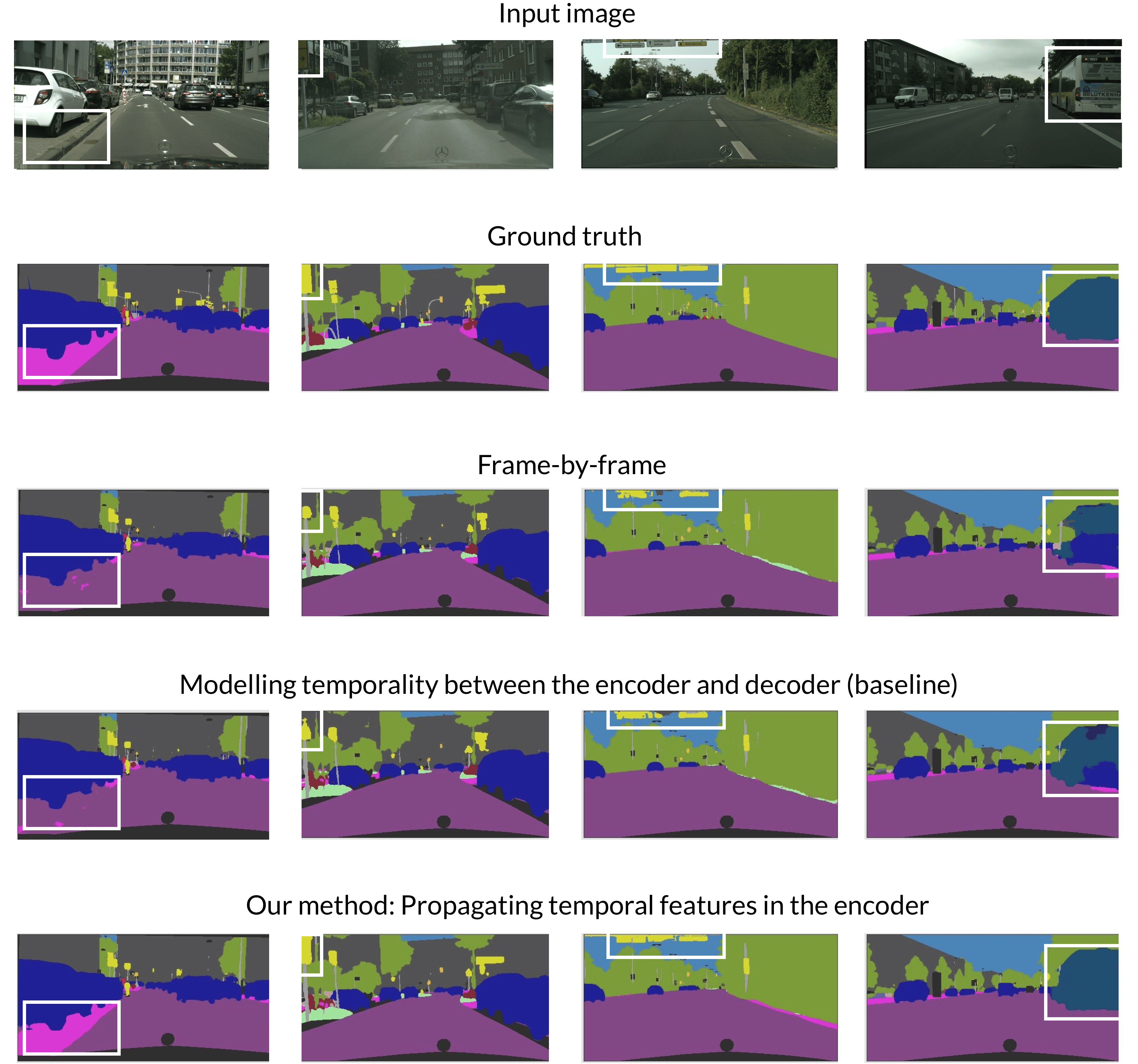}
\end{center}
   \caption{Qualitative results on the CityScapes validation set of the effect of different temporal module locations. White boxes highlight differences. We have used Pointwise TNs as a temporal module in this example.}
\label{where-to-model}
\end{figure*}

\begin{figure*}
\begin{center}
\includegraphics[width=\linewidth]{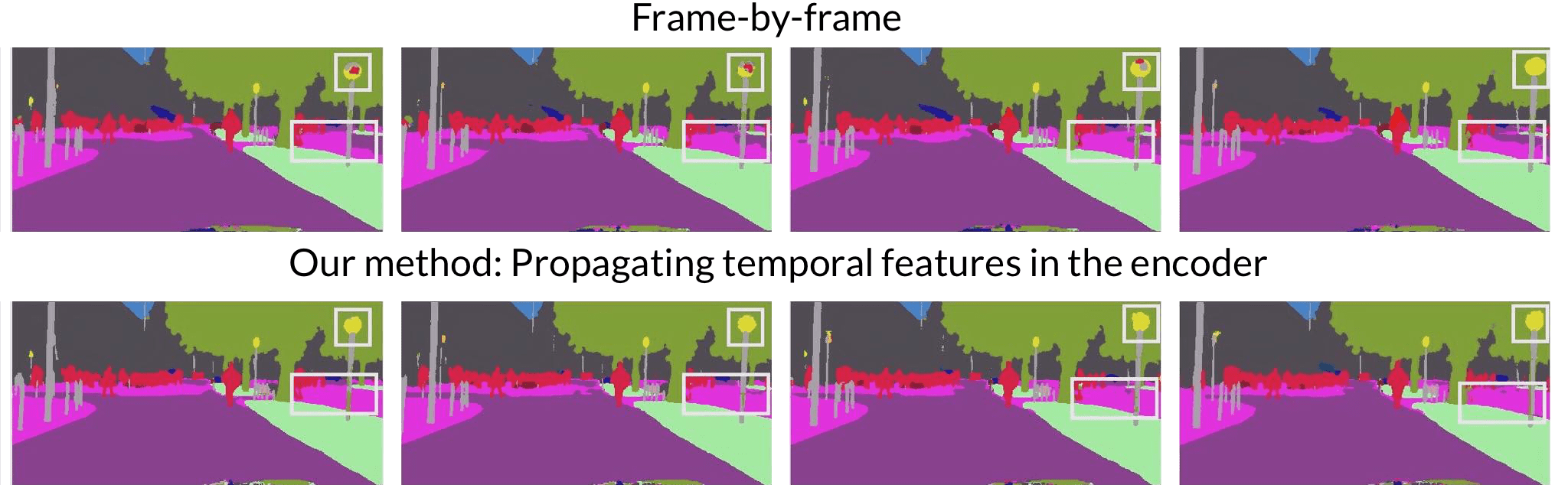}
\end{center}
   \caption{Examples of semantic segmentations in consecutive video frames from the CityScapes demo video. White boxes highlight differences. We have used Pointwise TNs as a temporal module in this example.}
\label{temporal-consistency}
\end{figure*}

\subsection{Quantitative evaluation} We compare the different temporal module locations and types using a U-Net architecture both qualitatively and quantitatively. In Table~\ref{results_ours}, we show that placing a temporal module between the encoder and decoder (baseline) increases the performance only by 1-2 percentage points, with any type of temporal module, when compared to the frame-by-frame baseline. Similarly, placing any type of temporal module at every skip connection level increases the mIoU by 3-4 percentage points. Most notably, also propagating these features to the next convolutional layers in the encoder results in a 5-6 percentage points performance increase. These results highlight that our temporal model significantly outperforms both the frame-by-frame image segmentation model and the baseline in which we only model temporality between the encoder and the decoder. Furthermore, we see that replacing ConvLSTMs with our Pointwise TNs only results in a 2 percentage point performance drop, while using only $T^2$ additional parameters instead of $4C^2K^2$. This results in a model with less than a third in size and thus 40\% faster training time in our implementation. In order to get results for the test set, we submitted our results to the CityScapes benchmark website. We have upsampled our predictions 4 times using nearest neighbour interpolation. The results can be seen in Table~\ref{results_test_ours} and further validate our hypothesis. We compare our results against other methods that perform inference at $256 \times 512$ resolution. Lastly, in order to check that the performance increase comes from the temporal information and not from the increased model complexity, we have sent the current frame 4 times as input for the temporal models. We have also concatenated the 4 consecutive images along the channel dimension, and sent it as input to the fully convolutional network. In both cases, the performance was similar or worse than the frame-by-frame segmentation model.

\subsection{Qualitative evaluation}

\par{\textbf{Temporal module locations.}}
In Figure~\ref{where-to-model}, we show several representative situations, in which our proposed approach to model temporality outperforms other methods. Observe how both the frame-by-frame and the baseline model struggle to segment objects that are mostly occluded in the current frame. Our poposed method is able to correct larger parts of inaccurate segmentation by leveraging the information available in the unlabeled temporal frames. We can conclude that only placing a temporal module between the encoder and decoder is suboptimal for capturing motion information.

\par
\begin{figure*}
\begin{center}
\includegraphics[width=\linewidth]{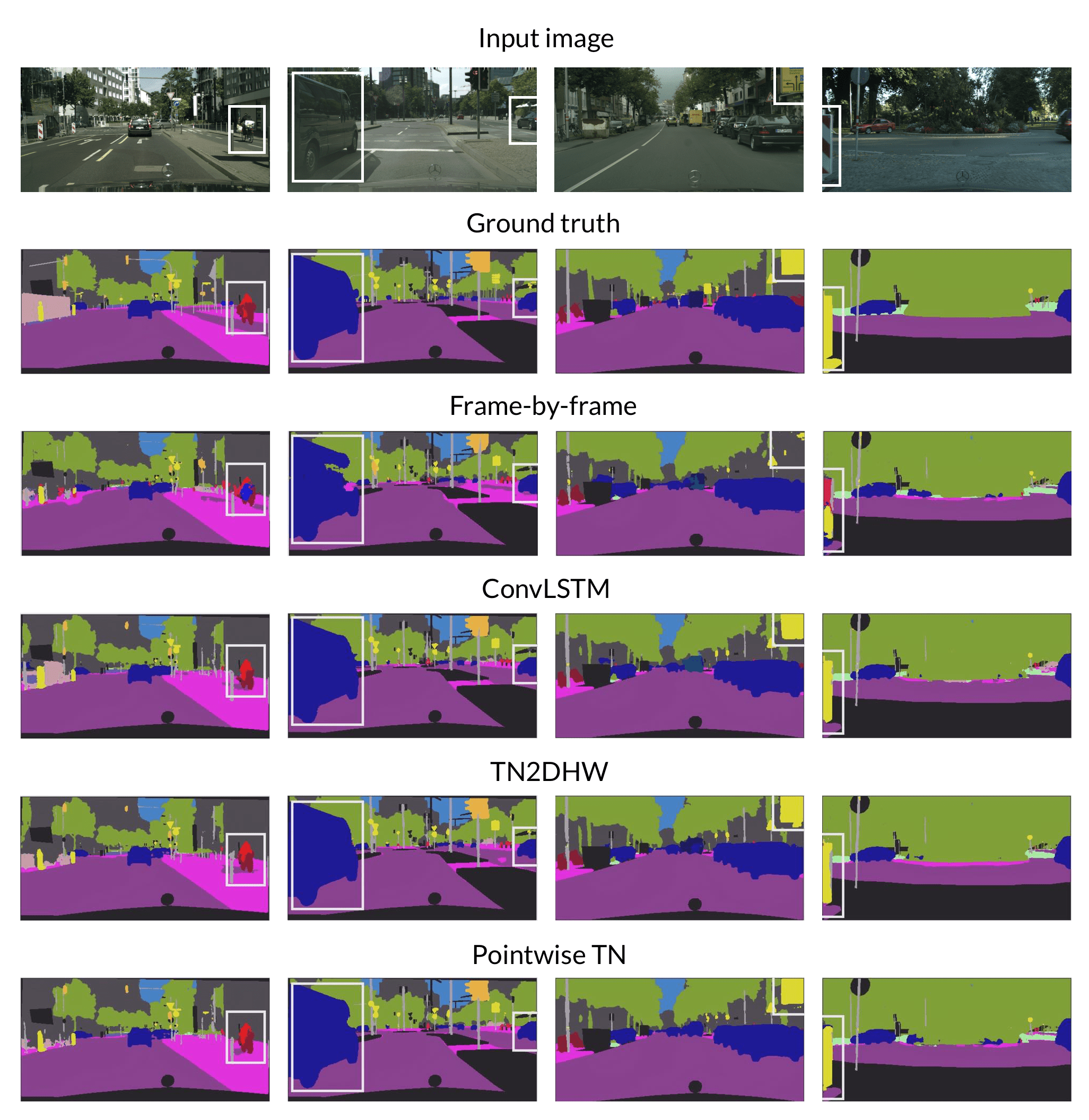}
\end{center}
   \caption{Qualitative results on the CityScapes validation set of the effect of different temporal modules for our proposed method. White boxes highlight differences.}
\label{propagating-information}
\end{figure*}

{\textbf{Temporal modules.}}
We qualitatively evaluate the performance of different temporal modules for our proposed method of propagating the temporal features in the encoder. The results can be seen in Figure~\ref{propagating-information}. Although ConvLSTMs show the best performance, both qualitatively and quantitatively, Pointwise TNs achieve almost the same performance, with much fewer parameters, resulting in lower memory requirements and faster training. We can see that all 3 temporal modules are able to correctly classify objects that cannot be accurately detected in the current frame by inferring them from previous frames.

\par{\textbf{Temporal consistency.}}
We also qualitatively evaluate the temporal consistency of our semantic video segmentation method. We use the 3 demo videos provided in the CityScapes dataset, that are 600, 1100 and 1200 frames long, respectively. For each output of our proposed method, the previous 4 frames were sent as input, resulting in a sliding window approach. Typical errors made by models that rely on single-frame estimates include partly segmented objects, temporal inconsistencies and flickering. In Figure~\ref{temporal-consistency}, we show that some of these errors are corrected when we are using the proposed temporal module. This behaviour is consistent across multiple sequences.

\section{Conclusions}
In this paper, we proposed a novel spatio-temporal neural network combining FCNs to model spatial information and temporal units to include temporal information. The proposed network was built on an encoder-decoder framework, which takes multiple continuous frames of a driving scene as input and outputs their semantic segmentation. We investigated different ways to model temporality in our network, as the temporal dimension is highly informative and can be critical in certain scenarios, such as autonomous driving. Starting with the conventional approach of placing a temporal module in between our encoder and decoder (baseline), we show that this method is suboptimal and that placing a temporal module after every encoder block results in much stronger propagation of temporal information and achieves significantly better results. In the qualitative results, we show that the proposed network is able to correctly classify objects that are mostly occluded in the current frame, by inferring them from the previous frames. This further illustrates the effectiveness of our method, as both the baseline and the frame-by-frame model are unable to correctly infer them. 

Furthermore, we experimented with different modules to model temporality. Although ConvLSTMs showed the best performance, stacking multiple instances of them in a network is costly, both in training speed and memory requirements. We tackle this issue by using Pointwise TNs, which achieve almost the same performance as the \mbox{ConvLSTM} while using a very low number parameters, which results in significantly lower memory requirements and faster \mbox{training}. 

The module can be easily added in state of the art methods, \eg DeepLab \cite{DBLP:journals/corr/abs-1802-02611}, by placing temporal modules after spatial convolutional modules. Preliminary experiments in this direction showed only minimal improvement. In \mbox{U-Net} architectures, temporality is injected in the final decoder layers through the use of skip connections. As newer models do not rely on skip connections, a solution is to add temporal modules in between convolutional layers in the decoder as well. Future work is needed to investigate the effectiveness of including temporality in such models. \href{https://github.com/mhashas/Exploiting-Temporality-For-Semi-Supervised-Video-Segmentation}{Code} has been made available.

{\small
\bibliographystyle{ieee}
\bibliography{egbib}
}

\end{document}